\documentclass[sigconf]{acmart}

\usepackage{cleveref}
\usepackage{enumitem}
\usepackage{multirow}

\usepackage{todonotes}
\usepackage{circledsteps}
\usepackage{xspace}
\usepackage{listings}

\makeatletter
\DeclareRobustCommand\onedot{\futurelet\@let@token\@onedot}
\def\@onedot{\ifx\@let@token.\else.\null\fi\xspace}

\makeatother

\AtBeginDocument{%
 \providecommand\BibTeX{{%
 \normalfont B\kern-0.5em{\scshape i\kern-0.25em b}\kern-0.8em\TeX}}}

\copyrightyear{2024}
\acmYear{2024}
\setcopyright{rightsretained}
\acmConference[GECCO '24 Companion]{Genetic and Evolutionary Computation Conference}{July 14--18, 2024}{Melbourne, VIC, Australia}
\acmBooktitle{Genetic and Evolutionary Computation Conference (GECCO '24 Companion), July 14--18, 2024, Melbourne, VIC, Australia}
\acmDOI{10.1145/3638530.3664163}
\acmISBN{979-8-4007-0495-6/24/07}

\begin{document}

\title[An investigation on the use of LLMs for hyperparameter tuning in EAs]{An investigation on the use of Large Language Models for hyperparameter tuning in Evolutionary Algorithms}


\author{Leonardo Lucio Custode}
\email{leonardo.custode@unitn.it}
\orcid{0000-0002-1652-1690}
\affiliation{%
 \institution{University of Trento}
 \city{Trento}
 \country{Italy}
}

\author{Fabio Caraffini}
\email{fabio.caraffini@swansea.ac.uk}
\orcid{0000-0001-9199-7368}
\affiliation{%
 \institution{Swansea University}
 \city{Swansea}
 \country{UK}
}

\author{Anil Yaman}
\email{a.yaman@vu.nl}
\orcid{0000-0003-1379-3778}
\affiliation{%
 \institution{Vrije Universiteit Amsterdam}
 \city{Amsterdam}
 \country{The Netherlands}
}

\author{Giovanni Iacca}
\email{giovanni.iacca@unitn.it}
\orcid{0000-0001-9723-1830}
\affiliation{%
 \institution{University of Trento}
 \city{Trento}
 \country{Italy}
}

\renewcommand{\shortauthors}{Custode et al.}

\begin{abstract}
Hyperparameter optimization is a crucial problem in Evolutionary Computation. In fact, the values of the hyperparameters directly impact the trajectory taken by the optimization process, and their choice requires extensive reasoning by human operators. Although a variety of self-adaptive Evolutionary Algorithms have been proposed in the literature, no definitive solution has been found. In this work, we perform a preliminary investigation to automate the reasoning process that leads to the choice of hyperparameter values. We employ two open-source Large Language Models (LLMs), namely Llama2-70b and Mixtral, to analyze the optimization logs online and provide novel real-time hyperparameter recommendations. We study our approach in the context of step-size adaptation for $(1+1)$-ES. The results suggest that LLMs can be an effective method for optimizing hyperparameters in Evolution Strategies, encouraging further research in this direction.
\end{abstract}

\ccsdesc[500]{Information systems~Language models}
\ccsdesc[500]{Theory of computation~Evolutionary algorithms}

\keywords{Evolutionary Algorithms, Large Language Models, Landscape Analysis, Parameter Tuning}

\maketitle

\section{Introduction}

It is well established in the field of Evolutionary Computation (EC), and even more broadly in Machine Learning (ML), that the hyperparameters of an algorithm significantly influence its performance. In evolutionary optimization, improper assignment of these parameters can lead to a suboptimal search process and provide inferior solutions \cite{yaman2019comparison}.
Over the past few decades, various approaches have been proposed to tackle this challenge by adjusting the parameters of the algorithms to positively influence the optimization process.
For instance, ontology-based approaches aim to formally represent the knowledge revolving around the EC field \cite{yaman2017presenting} to enable inference-based tuning, whereas ML-based approaches \cite{seiler2024deepela} learn useful representations for comparing evolutionary runs for landscape analysis. Other approaches pre-define pools of parameter values \cite{iacca2014differential,iacca2015continuous} and automatically choose from them during the optimization process. More recently, Reinforcement Learning (RL) has been considered to automatically train hyperparameter adaptation policies based on data from previously available evolutionary runs \cite{tessari2022reinforcement}. 
In general, all these approaches aim to find the optimal parameter settings of an Evolutionary Algorithm (EA) \emph{before} or \emph{during} the evolutionary optimization process and are usually referred to as \emph{parameter tuning} or \emph{parameter control}, respectively \cite{eiben1999parameter}.

Using data acquired from evolutionary runs is certainly a viable approach to leverage the knowledge acquired from past optimization experiences, as shown in \cite{tessari2022reinforcement}. However, the data generated by evolutionary runs can be highly unstructured, and they heavily depend on multiple aspects, such as the problem dimensionalities, the different algorithms and strategies being used, the numerical precision, the verbosity of the logging system in use, etc. 

In recent years, Large Language Models (LLMs) have caused a profound revolution in various fields, going beyond the traditional domain of conversational agents. To date, LLMs have been successfully applied to an ever-growing list of domains, such as chemistry \cite{boiko2023autonomous}, protein design \cite{ferruz2022controllable}, robotics \cite{wang2023gensim,ren2023robots,sun2023interactive}, swarm motion planning \cite{jiao2023swarm}, program synthesis \cite{10.1145/3510003.3510203,tao2023program,hemberg2024evolving}, game design \cite{lanzi2023chatgpt}, and urban delivery route optimization \cite{liu2023can}.

One of the main keys to the success of LLMs is that they are extremely well-suited for analyzing unstructured textual data, which makes it easy to adapt them to conceptually similar tasks without the need for retraining.
Some recent literature \cite{yang2023large,guo2023connecting} has investigated the direct usage of LLMs for optimization, although it is important to note that current LLMs are well-suited especially for high-level discrete optimization tasks (e.g., planning).
On the other hand, they are not well-suited for low-level (i.e., numerical) continuous optimization tasks due to (1) the inference cost of querying LLMs, and (2) their limited mathematical reasoning capabilities.
Finally, several works have investigated the relationship between LLM and EC \cite{wu2024evolutionary}, both in the cases where LLMs are used to empower EAs, and in the opposite case.
However, none of them studied the use case of hyperparameter optimization, which is an essential problem in EC.

In this work, we conduct a preliminary investigation on the possibility of automating ``empirical'' step-size adaptation in Evolution Strategies (ES) using LLMs.
More specifically, we focus on the case of tuning the step-size for $(1+1)$-ES using state-of-the-art LLMs, namely Llama2-70b \cite{touvron2023llama} and Mixtral \cite{jiang2024mixtral}.
Our results show that LLMs can compete with well-known traditional mechanisms for step-size adaptation.

The remainder of the paper is organized as follows. \Cref{sec:rw} summarizes the related work on LLMs and their applications to optimization, as well as the early work on EC applied to LLMs. \Cref{sec:methods} describes our methods. \Cref{sec:results} presents the numerical results. Finally, \Cref{sec:concl} concludes this research.


\begin{figure*}
 \centering
 \includegraphics[width=0.6\textwidth]{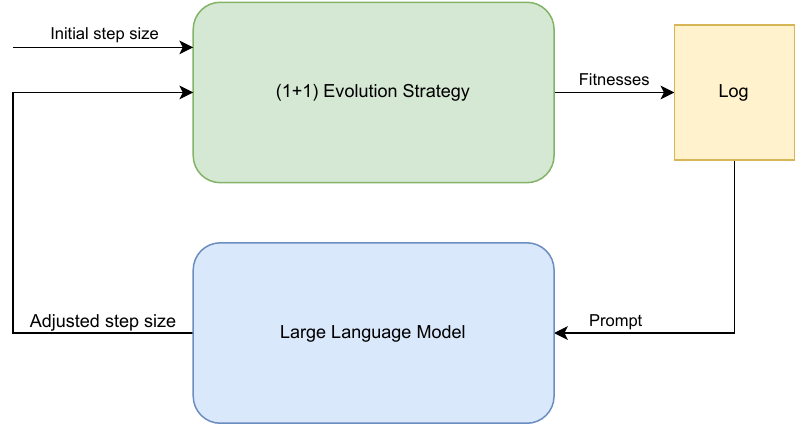}
 \caption{A block diagram showing the self-adaptive step-size adaptation scheme studied in this work.}
 \Description{}
 \label{fig:block_diagram}
\end{figure*}

\section{Related work}
\label{sec:rw}

There is a growing trend to combine the strengths of optimization algorithms with LLMs for various purposes. This synergy is evident also in the EC research community, where the `LLM' keyword is increasingly prevalent and the increasing number of available \textit{preprints}, see \cite{wu2024evolutionary} for a recent exhaustive list, is a clear signal that more articles of this kind will soon populate the literature. This highlights a promising direction where LLMs will play an important role in optimization (especially, but not only, evolutionary-based), addressing various challenges associated with optimization algorithms and leading to hybrid systems where optimization algorithms and LLMs work together for mutual benefits.

\subsection{LLMs as the optimizer}
LLMs make it possible to articulate complex tasks using natural language, thus facilitating and automating decision-making in many logistical scenarios, such as planning and scheduling \cite{pallagani2024prospects}, which otherwise would require an expert to formulate the problem rigorously (i.e., as a discrete optimization task). For example, ChatGPT has been used to produce the schedule of a construction project in \cite{buildings13040857} and to schedule vehicles and drivers to complete a predetermined set of trips in a specified period in \cite{chatGPTInLogistics}. These are combinatorial optimization problems whose operational constraints are difficult to formulate mathematically but easy to communicate to an LLM. Similarly, in \cite{li2023large}, users can ask ``what if'' questions on supply chain demand for a particular scenario in plain text and get answers about the outcomes of an underlying supply chain optimization algorithm.

Hyperparameter optimization is another intriguing application where LLMs can act as the optimizer for other ML models. Compared to traditional hyperparameter optimization methods, the results appear promising \cite{zhang2023using}. For instance, the studies in \cite{zheng2023gpt4} and \cite{zhang2023automlgpt} leverage GPT-4 to perform Neural Architecture Search and design autoML frameworks, respectively.

Finally, Optimization by PROmpting (OPRO) is a framework proposed in \cite{yang2023large} for using LLMs as an iterative heuristic for gradient-free optimization. In each optimization step, the OPRO LLM generates new solutions from those contained (including their values) in the previous prompt. As with any EC approach, novel candidate solutions are evaluated and added to the next prompt for the next optimization step. This is a general approach that is usable for both discrete and continuous problems. The results presented in \cite{yang2023large} for linear regression and Travelling Salesman Problems are promising, highlighting that OPRO can optimize the prompts without human intervention.

\subsection{EAs for prompt or LLM optimization}

The need for a good quality prompt has opened the way for the prompt optimization/engineering field. The existing literature explores different methodologies to optimize prompts.
Recent work \cite{pryzant2023automatic} suggests using gradient descent with beam search and bandit selection assuming that training data and an LLM API are accessible. EC offers various tools for this task, which can also be used when training data and internal LLM information are inaccessible (i.e., the gradient information cannot be exploited). Genetic Algorithms are widely used \cite{feng2024genetic,ijcai2023p588}, and ideas borrowed from EAs are explored in the EvoPrompt framework \cite{guo2024connecting}. In \cite{pmlr-v162-sun22e}, an interesting approach based on vectorizing prompts in a continuous subspace employs CMA-ES to perform the optimization process.

LLM architecture search is another intriguing optimization challenge. EAs can greatly help explore these vast search spaces. In the AutoBERT-Zero framework \cite{gao2022autobert}, this is done by performing primitive math operations and through an interesting ``Operation-Priority'' evolution strategy, which exploits prior information from operations during the search. Other studies consider multiple objectives, such as the one in \cite{tribes2024hyperparameter}, which compares direct search methods and classic heuristics, or the framework in \cite{ganesan2021supershaper}, where EAs are used to find the most suitable distribution of hidden units across layers, ensuring that accuracy and latency requirements are met concurrently.

\subsection{LLMs supporting EAs}

LLMs can support EAs in numerous ways and there is growing enthusiasm for employing them to solve various mathematical problems. In this context, the most popular example of incorporating LLM into EAs is probably the FunSearch algorithm \cite{romera2023mathematical}. This algorithm has raised a debate among researchers, with some applauding its ability to tackle unresolved mathematical problems and demonstrate genuine ``intelligence'', while others pointing out its limitations, as being ``\textit{remarkable for the shallowness of the mathematical understanding that it seems to exhibit}'' \cite{ernestdavies}. Regardless of its public perception, FunSearch is an interesting, but algorithmically straightforward approach. It involves employing an LLM to generate code for a specific subroutine within a larger program in a Genetic Programming (GP) algorithm, meaning that the LLM is embedded into the EA to perform the mutation step. The idea of using LLMs for code generation in EAs is not new. In \cite{lehman2023evolution}, they are used with the MAP-Elites algorithm to generate effective mutation operators for GP.

Algorithm design is another application of LLMs to EAs. For example, in \cite{liu2023algorithm} LLMs are employed to generate and evolve algorithm components such as initialization, selection, crossover, etc. to create new algorithms.

There are interesting uses for LLMs in other optimization scenarios. The system described in \cite{amarasinghe2023aicopilot} uses these techniques to simplify the formulation of common complex business optimization problems, allowing the optimizer to improve performance in production scheduling. In \cite{LIU2021310}, LLMs check the feasibility of solutions to constraints in discrete spaces to facilitate the application of simulated annealing and other single-solution metaheuristics. LLMs can also function as supervisory logic, e.g., they coordinate agents in the coevolutionary algorithm presented in \cite{electronics12122722}.


\section{Methods}
\label{sec:methods}
In this work, we leverage LLMs to automatically optimize the step-size for $(1+1)$-ES, which generates an offspring by adding, to the current solution, a uniform perturbation in a hyper-sphere of radius $\sigma$, the step-size.

In our method, the LLM is queried every $p$ optimization steps (with $p$ being user-defined, e.g. based on the available computational resources) with the log produced from the beginning of the optimization process to the current generation. The LLM is then asked to produce a recommendation for the next step-size to be used in the subsequent $p$ steps.
For the LLM-based step-size adaptation, we use a period of $p=50$ generations to calculate the new step-size.

We fix the number of fitness evaluations for all the methods to $10^3$.
Our code is publicly available on GitHub\footnote{\href{https://github.com/DIOL-UniTN/llm_step_size_adaptation}{https://github.com/DIOL-UniTN/llm\_step\_size\_adaptation}}.

\subsection{Benchmark problems}
We run $(1+1)$-ES on the BBOB suite \cite{finck2010real}, using the IOHProfiler suite \cite{IOHprofiler}.
We test all the BBOB problems with dimensionalities $2$, $5$, and $30$, to assess the capabilities of the proposed method to work in both low- and moderate-sized problems. 
For each function, we perform $10$ independent runs with different starting points. Then, we feed the logs to the LLMs, using the prompt described below.

\subsection{Prompt format}
To provide the LLM with the information needed to tune the step-size, we initially considered three different options: (1) feeding only the fitness values; (2) feeding genotypes and fitnesses; and (3) feeding the genotypes, the relationships between the genotypes (i.e., between the parent solution and its offspring, through the effect of mutation), and the fitnesses.
However, after preliminary experiments (not reported here for brevity), we did notice that the last two approaches required a substantially higher amount of tokens, quickly filling the context window of the models.
For this reason, we eventually use a prompt that only contained the essential information to make a decision, i.e., the changes in the step-size and the fitnesses of all the last individuals evaluated (depending on the size of the context window).

An example of such a prompt can be seen in Listing \ref{list:prompt}.
Here, the ``system'' prompt is used to provide a preliminary context to the model to properly address the task. Then, the ``user'' task represents the actual query.
Finally, it is important to note that we request a step-size $s \in [0.001, 0.999]$ to prevent the LLMs from choosing exploding step-sizes.

\subsection{Algorithms and LLMs}
Our comparison investigates the performance of different strategies for adapting the step-size of $(1+1)$-ES.
We employ two baseline methods and two LLM-based methods.
The two baselines consist of: (1) a constant step-size for the ES (fixed to $0.1$); and (2) a well-known rule-based strategy, namely the One-Fifth success rule \cite{rechenberg1973evolutionsstrategie}. The latter changes the step-size by considering the current candidate $x' = x + \mathcal{N}(0, \sigma_s)$, where $x$ is the parent solution, $\mathcal{N}$ is the Gaussian (normal) distribution, and $\sigma_s$ is the step-size, as follows:
\begin{equation}
 \sigma'_s = \begin{cases}
 1.5 \sigma_s & if~f(x') < f(x) \\
 1.5^{-\frac{1}{4}} \sigma_s & if~f(x') > f(x) \\
 \sigma_s & if~f(x') = f(x) \\
 \end{cases}
\end{equation}

Regarding the LLM-based step-size adaptation strategies, we employ two widely known open-source state-of-the-art LLMs, namely Llama2-70b and Mixtral.
Llama2-70b is the largest model of the Llama2 family \cite{touvron2023llama}. It features $80$ transformer layers with a hidden size of $8192$ dimensions \cite{chen2024bigger}. It has been trained on $2T$ tokens and has a maximum context size of $4096$ tokens.

Mixtral \cite{jiang2024mixtral}, instead, is a mixture-of-experts model made of $8$ experts, each with $32$ transformer layers with a hidden size of $4096$ dimensions. It has a context size of $32k$ tokens.

For both models, we employ Groq's APIs for accelerated inference\footnote{https://groq.com}.

\begin{table*}
\begin{lstlisting}[frame=single,caption=Prompt template.,label=list:prompt]
{
 "role": "system",
 "content": "You are a powerful and intelligent AI capable of analyzing logs and
 performing reasoning"
},
{
 "role": "user",
 "content": "Q: I am running an optimization process over an unknown function.
 I am using a 1+1-ES to optimize the function.
 f(x) = y indicates an evaluation.
 x1 -> x2 indicates that x1, using the current step size,
 produced a new candidate solution x2.
 It is extremely important that the step size you propose is
 contained between 0.999 and 0.001.
 Here's the log:```txt
 <LOG CONTENT>
 ```
 I am currently using the following step size: <STEP SIZE>.
 Should I change it or not?
 Do you think that the current step size is good enough to make
 the process converge as soon as possible?
 Reply with the following structure:
 `Reasoning: <explanation>
 Recommended step size: <new step size>`"
}
\end{lstlisting}
\end{table*}

\section{Results}
\label{sec:results}

\begin{figure*}[ht!]
 \centering
 \includegraphics[width=\textwidth]{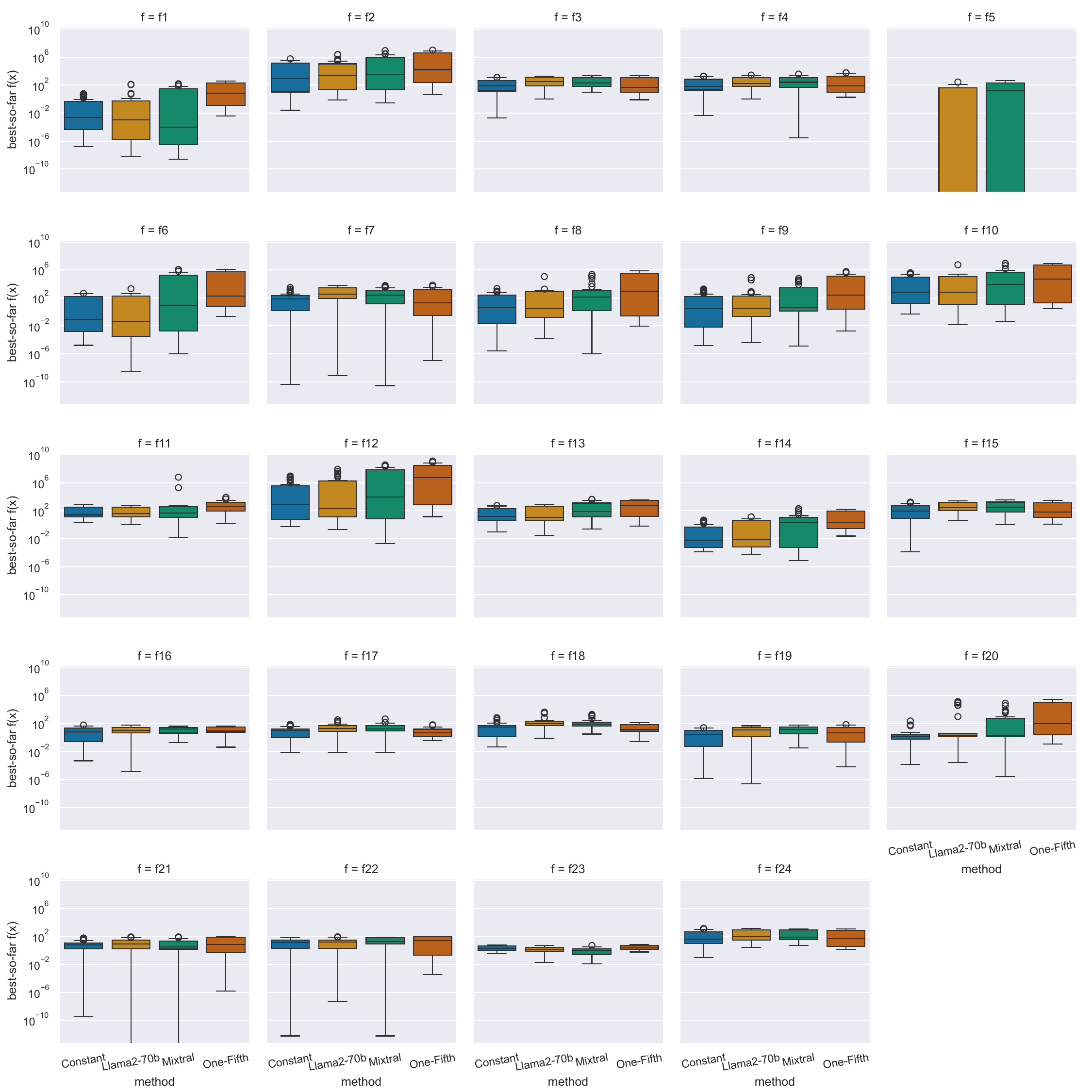}
 \caption{Boxplot of the best fitness obtained in 10 runs by each of the methods on each function from the BBOB suite. Note that we aggregate the values from $10$ runs on all the considered problem sizes.}
 \Description{}
 \label{fig:boxplot}
\end{figure*}

The results obtained by the $4$ methods under comparison are shown in \Cref{fig:boxplot}, aggregating the results for all the problem sizes to facilitate visualization.
We observe that, in most cases, the performance of both Llama2-70b and Mixtral is comparable to that of the One-Fifth strategy, suggesting a somehow similar behavior in terms of step-size adaptation.
Note that, for the $f5$ function, the methods using constant step-sizes and the one-fifth rule achieve values very close to zero, and thus they are not visible in the boxplot (which uses a logarithmic scale).

\begin{figure*}[t!]
 \centering
 \includegraphics[width=\textwidth]{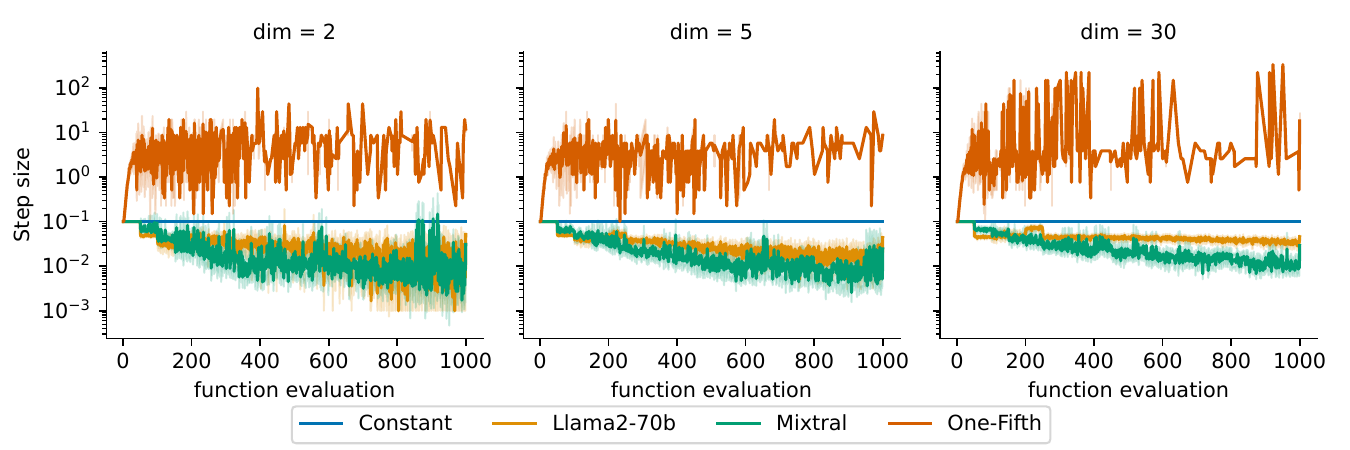}
 \caption{Average step-size for each function evaluation for all the methods under comparison. The solid lines represent the mean value for the step-size (averaged over all problems and problem dimensionalities), while the shaded area represents the 95\% confidence interval.}
 \Description{}
 \label{fig:stepsizeplot}
\end{figure*}

To verify this hypothesis, we studied the step-size used by each of the methods under comparison in \Cref{fig:stepsizeplot}.
Note that the trend shown in \Cref{fig:stepsizeplot} contains the average step-size computed on all the functions of the benchmark, divided by all the dimensionalities considered before.
We observe that, contrary to our expectations, their impact on the step-size is actually at two opposite extremes.
In fact, while the One-Fifth rule pushes the step-size to very high values, the two LLM-based methods tend to progressively reduce the step-size. Yet, their performance appears to be overall similar.
Moreover, it is interesting to note that the variance of the LLM-based step-size adaptation techniques is inversely proportional to the problem dimensionality, while the variance of the one-fifth adaptation techniques seems much larger in the highest-dimensionality case.

Finally, to add a further level of comparative evaluation of the four methods, we performed the glicko2-based ranking \cite{glickman2012example} for fixed-budget scenarios, available in the IOHAnalyzer platform\footnote{https://iohanalyzer.liacs.nl/}. The results are shown in \Cref{tab:glicko_d2,tab:glicko_d5,tab:glicko_d30}.
In the tables, we observe an interesting trend: Llama2-70b always ranks first, outperforming all the other methods.
On the other hand, we see Mixtral as a runner-up when the problem dimensionality is $2$, and it gradually ranks worse as the dimensionality of the problem increases.
Another interesting remark is that, in the highest-dimensionality case (\Cref{tab:glicko_d30}), both Mixtral and the One-Fifth rule rank worse than having a constant step-size.
On the other hand, in 30 dimensions Llama2-70b is the only method that performs better than simply using a constant step-size.

These preliminary results indicate that it is indeed possible to have well-performing step-size adaptation strategies by using the reasoning capabilities in LLMs. However, our results also show that the outcomes are heavily dependent on the LLM at hand, suggesting that a specialized LLM (e.g., fine-tuned on a large dataset of evolutionary runs) could lead to even better results.

\begin{table*}[ht!]
 \centering
 \caption{Glicko2-based ranking of the methods on functions from the BBOB suite with dimensionality 2.}
 \label{tab:glicko_d2}
 \begin{tabular}{lcccccccc}
 \toprule
 \textbf{Step-size adaptation method} & \textbf{Rating} & \textbf{Deviation} & \textbf{Volatility} & \textbf{Games} & \textbf{Win} & \textbf{Draw} & \textbf{Loss}\\
 \midrule
 Llama2-70b & 1532 & 18.9 & 0.0480 & 5700 & 3189 & 9 & 2502 \\
 Mixtral & 1510 & 18.0 & 0.0431 & 5700 & 2853 & 9 & 2838 \\
 Constant step-size & 1479 & 17.5 & 0.0409 & 5700 & 2705 & 6 & 2989 \\
 One-Fifth rule & 1479 & 17.3 & 0.0392 & 5700 & 2640 & 2 & 3058 \\
 \bottomrule
 \end{tabular}
\end{table*}

\begin{table*}[ht!]
 \caption{Glicko2-based ranking of the methods on functions from the BBOB suite with dimensionality 5.}
 \label{tab:glicko_d5}
 \begin{tabular}{lcccccccc}
 \toprule
 \textbf{Step-size adaptation method} & \textbf{Rating} & \textbf{Deviation} & \textbf{Volatility} & \textbf{Games} & \textbf{Win} & \textbf{Draw} & \textbf{Loss} \\
 \midrule
 Llama2-70b & 1562 & 17.8 & 0.0415 & 5700 & 3537 & 0 & 2163 \\
 One-Fifth rule & 1495 & 16.6 & 0.0365 & 5700 & 2778 & 0 & 2922 \\
 Mixtral & 1481 & 18.1 & 0.0436 & 5700 & 2695 & 0 & 3005 \\
 Constant step-size & 1459 & 16.7 & 0.0359 & 5700 & 2390 & 0 & 3310 \\
\bottomrule
\end{tabular}
\end{table*}

\begin{table*}[ht!]
 \caption{Glicko2-based ranking of the methods on functions from the BBOB suite with dimensionality 30.}
 \label{tab:glicko_d30}
 \begin{tabular}{lcccccccc}
 \toprule
 \textbf{Step-size adaptation method} & \textbf{Rating} & \textbf{Deviation} &  \textbf{Volatility} & \textbf{Games} & \textbf{Win} & \textbf{Draw} & \textbf{Loss} \\
 \midrule
 Llama2-70b & 1570 & 17.3 & 0.0378 & 5700 & 3591 & 0 & 2109 \\
 Constant step-size & 1540 & 16.6 & 0.0354 & 5700 & 3202 & 0 & 2498\\
 One-Fifth rule & 1457 & 16.7 & 0.0361 &5700 & 2363 & 0 & 3337 \\
 Mixtral & 1430 & 17.5 &0.0390 & 5700 & 2244 & 0 & 3456\\
\bottomrule
\end{tabular}
\end{table*}


\section{Conclusions}
\label{sec:concl}
The step-size is a crucial parameter in ES.
In fact, it allows the user to directly control the exploration and exploitation capabilities of the algorithm.
While several approaches have been proposed for adapting the step-size in ES, there is no one-fits-all strategy that solves this problem.
This is due to the fact that step-size control heavily depends on the context at hand, which requires some form of reasoning.
In this direction, we performed a preliminary investigation on the use of LLMs for tuning the step-size of $(1+1)$-ES. Our results indicate that it is possible to find adaptation strategies that are competitive with established methods for step-size adaptation, such as the One-Fifth rule.
Moreover, given the fixed budget used in our experiments, the approach based on Llama2-70b showed better consistency w.r.t. the other techniques, always ranking as the best algorithm.

The present work suggests several interesting future directions, such as experimenting with different prompts (and the related information therein) provided to the LLMs, fine-tuning existing open-source LLMs to the specific domain of adaptation strategies, conducting a more extensive study on hyperparameter optimization for EAs using LLMs (i.e., using larger budgets, as suggested in \cite{auger2009benchmarking}, and more sophisticated step-size adaptation schemes), and finally developing an LLM-guided EA.


\balance
\bibliographystyle{ACM-Reference-Format}
\bibliography{biblio} 


\begin{thebibliography}{51}


\ifx \showCODEN    \undefined \def \showCODEN     #1{\unskip}     \fi
\ifx \showDOI      \undefined \def \showDOI       #1{#1}\fi
\ifx \showISBNx    \undefined \def \showISBNx     #1{\unskip}     \fi
\ifx \showISBNxiii \undefined \def \showISBNxiii  #1{\unskip}     \fi
\ifx \showISSN     \undefined \def \showISSN      #1{\unskip}     \fi
\ifx \showLCCN     \undefined \def \showLCCN      #1{\unskip}     \fi
\ifx \shownote     \undefined \def \shownote      #1{#1}          \fi
\ifx \showarticletitle \undefined \def \showarticletitle #1{#1}   \fi
\ifx \showURL      \undefined \def \showURL       {\relax}        \fi
\providecommand\bibfield[2]{#2}
\providecommand\bibinfo[2]{#2}
\providecommand\natexlab[1]{#1}
\providecommand\showeprint[2][]{arXiv:#2}

\bibitem[\protect\citeauthoryear{Auger}{Auger}{2009}]%
        {auger2009benchmarking}
\bibfield{author}{\bibinfo{person}{Anne Auger}.} \bibinfo{year}{2009}\natexlab{}.
\newblock \showarticletitle{{Benchmarking the (1+1) evolution strategy with one-fifth success rule on the BBOB-2009 function testbed}}. In \bibinfo{booktitle}{\emph{Genetic and Evolutionary Computation Conference Companion}}. \bibinfo{publisher}{ACM}, \bibinfo{address}{{New York, NY, USA}}, \bibinfo{pages}{2447--2452}.
\newblock


\bibitem[\protect\citeauthoryear{{Beibin Li and Konstantina Mellou and Bo Zhang and Jeevan Pathuri and Ishai Menache}}{{Beibin Li and Konstantina Mellou and Bo Zhang and Jeevan Pathuri and Ishai Menache}}{2023}]%
        {li2023large}
\bibfield{author}{\bibinfo{person}{{Beibin Li and Konstantina Mellou and Bo Zhang and Jeevan Pathuri and Ishai Menache}}.} \bibinfo{year}{{2023}}\natexlab{}.
\newblock \bibinfo{title}{{Large Language Models for Supply Chain Optimization}}.
\newblock
\newblock
\newblock
\shownote{{arXiv:2307.03875}}.


\bibitem[\protect\citeauthoryear{{Boiko, Daniil A and MacKnight, Robert and Kline, Ben and Gomes, Gabe}}{{Boiko, Daniil A and MacKnight, Robert and Kline, Ben and Gomes, Gabe}}{2023}]%
        {boiko2023autonomous}
\bibfield{author}{\bibinfo{person}{{Boiko, Daniil A and MacKnight, Robert and Kline, Ben and Gomes, Gabe}}.} \bibinfo{year}{{2023}}\natexlab{}.
\newblock \showarticletitle{{Autonomous chemical research with large language models}}.
\newblock \bibinfo{journal}{\emph{{Nature}}} \bibinfo{volume}{{624}}, \bibinfo{number}{{7992}} (\bibinfo{year}{{2023}}), \bibinfo{pages}{{570--578}}.
\newblock


\bibitem[\protect\citeauthoryear{Chen, Wu, Liang, Gong, Shou, Zhang, and Li}{Chen et~al\mbox{.}}{2024}]%
        {chen2024bigger}
\bibfield{author}{\bibinfo{person}{Nuo Chen}, \bibinfo{person}{Ning Wu}, \bibinfo{person}{Shining Liang}, \bibinfo{person}{Ming Gong}, \bibinfo{person}{Linjun Shou}, \bibinfo{person}{Dongmei Zhang}, {and} \bibinfo{person}{Jia Li}.} \bibinfo{year}{2024}\natexlab{}.
\newblock \bibinfo{title}{{Is Bigger and Deeper Always Better? Probing LLaMA Across Scales and Layers}}.
\newblock
\newblock
\newblock
\shownote{arXiv:2312.04333}.


\bibitem[\protect\citeauthoryear{{Chengrun Yang and Xuezhi Wang and Yifeng Lu and Hanxiao Liu and Quoc V. Le and Denny Zhou and Xinyun Chen}}{{Chengrun Yang and Xuezhi Wang and Yifeng Lu and Hanxiao Liu and Quoc V. Le and Denny Zhou and Xinyun Chen}}{2023}]%
        {yang2023large}
\bibfield{author}{\bibinfo{person}{{Chengrun Yang and Xuezhi Wang and Yifeng Lu and Hanxiao Liu and Quoc V. Le and Denny Zhou and Xinyun Chen}}.} \bibinfo{year}{{2023}}\natexlab{}.
\newblock \bibinfo{title}{{Large Language Models as Optimizers}}.
\newblock
\newblock
\newblock
\shownote{{arXiv:2309.03409}}.


\bibitem[\protect\citeauthoryear{Davis}{Davis}{2024}]%
        {ernestdavies}
\bibfield{author}{\bibinfo{person}{Ernest Davis}.} \bibinfo{year}{2024}\natexlab{}.
\newblock \bibinfo{title}{{Using a large language model to generate program mutations for a genetic algorithm to search for solutions to combinatorial problems: Review of (Romera-Paredes et al., 2023).}}
\newblock
\newblock
\urldef\tempurl%
\url{https://cs.nyu.edu/~davise/papers/FunSearch.pdf}
\showURL{%
\tempurl}
\newblock
\shownote{Accessed on 7/04/2024}.


\bibitem[\protect\citeauthoryear{{de Zarzà, I. and de Curtò, J. and Roig, Gemma and Manzoni, Pietro and Calafate, Carlos T.}}{{de Zarzà, I. and de Curtò, J. and Roig, Gemma and Manzoni, Pietro and Calafate, Carlos T.}}{2023}]%
        {electronics12122722}
\bibfield{author}{\bibinfo{person}{{de Zarzà, I. and de Curtò, J. and Roig, Gemma and Manzoni, Pietro and Calafate, Carlos T.}}} \bibinfo{year}{{2023}}\natexlab{}.
\newblock \showarticletitle{{Emergent Cooperation and Strategy Adaptation in Multi-Agent Systems: An Extended Coevolutionary Theory with LLMs}}.
\newblock \bibinfo{journal}{\emph{{Electronics}}} \bibinfo{volume}{{12}}, \bibinfo{number}{{12}} (\bibinfo{year}{{2023}}), \bibinfo{pages}{{19}}.
\newblock
\showISSN{{2079-9292}}


\bibitem[\protect\citeauthoryear{Doerr, Wang, Ye, van Rijn, and B{\"a}ck}{Doerr et~al\mbox{.}}{2018}]%
        {IOHprofiler}
\bibfield{author}{\bibinfo{person}{Carola Doerr}, \bibinfo{person}{Hao Wang}, \bibinfo{person}{Furong Ye}, \bibinfo{person}{Sander van Rijn}, {and} \bibinfo{person}{Thomas B{\"a}ck}.} \bibinfo{year}{2018}\natexlab{}.
\newblock \bibinfo{title}{{IOHprofiler: A Benchmarking and Profiling Tool for Iterative Optimization Heuristics}}.
\newblock
\newblock
\newblock
\shownote{arXiv:1810.05281}.


\bibitem[\protect\citeauthoryear{{Eiben, {\'A}goston E and Hinterding, Robert and Michalewicz, Zbigniew}}{{Eiben, {\'A}goston E and Hinterding, Robert and Michalewicz, Zbigniew}}{1999}]%
        {eiben1999parameter}
\bibfield{author}{\bibinfo{person}{{Eiben, {\'A}goston E and Hinterding, Robert and Michalewicz, Zbigniew}}.} \bibinfo{year}{{1999}}\natexlab{}.
\newblock \showarticletitle{{Parameter control in evolutionary algorithms}}.
\newblock \bibinfo{journal}{\emph{{IEEE Transactions on evolutionary computation}}} \bibinfo{volume}{{3}}, \bibinfo{number}{{2}} (\bibinfo{year}{{1999}}), \bibinfo{pages}{{124--141}}.
\newblock


\bibitem[\protect\citeauthoryear{{Fei Liu and Xialiang Tong and Mingxuan Yuan and Qingfu Zhang}}{{Fei Liu and Xialiang Tong and Mingxuan Yuan and Qingfu Zhang}}{2023}]%
        {liu2023algorithm}
\bibfield{author}{\bibinfo{person}{{Fei Liu and Xialiang Tong and Mingxuan Yuan and Qingfu Zhang}}.} \bibinfo{year}{{2023}}\natexlab{}.
\newblock \bibinfo{title}{{Algorithm Evolution Using Large Language Model}}.
\newblock
\newblock
\newblock
\shownote{{arXiv:2311.15249}}.


\bibitem[\protect\citeauthoryear{Feng, Sun, Li, Zhou, Lv, and Lu}{Feng et~al\mbox{.}}{2024}]%
        {feng2024genetic}
\bibfield{author}{\bibinfo{person}{Chengzhe Feng}, \bibinfo{person}{Yanan Sun}, \bibinfo{person}{Ke Li}, \bibinfo{person}{Pan Zhou}, \bibinfo{person}{Jiancheng Lv}, {and} \bibinfo{person}{Aojun Lu}.} \bibinfo{year}{2024}\natexlab{}.
\newblock \bibinfo{title}{{Genetic Auto-prompt Learning for Pre-trained Code Intelligence Language Models}}.
\newblock
\newblock
\newblock
\shownote{arXiv:2403.13588}.


\bibitem[\protect\citeauthoryear{{Ferruz, Noelia and H{\"o}cker, Birte}}{{Ferruz, Noelia and H{\"o}cker, Birte}}{2022}]%
        {ferruz2022controllable}
\bibfield{author}{\bibinfo{person}{{Ferruz, Noelia and H{\"o}cker, Birte}}.} \bibinfo{year}{{2022}}\natexlab{}.
\newblock \showarticletitle{{Controllable protein design with language models}}.
\newblock \bibinfo{journal}{\emph{{Nature Machine Intelligence}}} \bibinfo{volume}{{4}}, \bibinfo{number}{{6}} (\bibinfo{year}{{2022}}), \bibinfo{pages}{{521--532}}.
\newblock


\bibitem[\protect\citeauthoryear{Finck, Hansen, Ros, and Auger}{Finck et~al\mbox{.}}{2010}]%
        {finck2010real}
\bibfield{author}{\bibinfo{person}{Steffen Finck}, \bibinfo{person}{Nikolaus Hansen}, \bibinfo{person}{Raymond Ros}, {and} \bibinfo{person}{Anne Auger}.} \bibinfo{year}{2010}\natexlab{}.
\newblock \bibinfo{booktitle}{\emph{{Real-parameter black-box optimization benchmarking 2009: Presentation of the noiseless functions}}}.
\newblock \bibinfo{type}{{T}echnical {R}eport}. \bibinfo{institution}{Citeseer}.
\newblock


\bibitem[\protect\citeauthoryear{Ganesan, Ramesh, and Kumar}{Ganesan et~al\mbox{.}}{2021}]%
        {ganesan2021supershaper}
\bibfield{author}{\bibinfo{person}{Vinod Ganesan}, \bibinfo{person}{Gowtham Ramesh}, {and} \bibinfo{person}{Pratyush Kumar}.} \bibinfo{year}{2021}\natexlab{}.
\newblock \bibinfo{title}{{SuperShaper: Task-Agnostic Super Pre-training of BERT Models with Variable Hidden Dimensions}}.
\newblock
\newblock
\newblock
\shownote{arXiv:2110.04711}.


\bibitem[\protect\citeauthoryear{Gao, Xu, Shi, Ren, Philip, Liang, Jiang, and Li}{Gao et~al\mbox{.}}{2022}]%
        {gao2022autobert}
\bibfield{author}{\bibinfo{person}{Jiahui Gao}, \bibinfo{person}{Hang Xu}, \bibinfo{person}{Han Shi}, \bibinfo{person}{Xiaozhe Ren}, \bibinfo{person}{LH Philip}, \bibinfo{person}{Xiaodan Liang}, \bibinfo{person}{Xin Jiang}, {and} \bibinfo{person}{Zhenguo Li}.} \bibinfo{year}{2022}\natexlab{}.
\newblock \showarticletitle{{AutoBERT-Zero: Evolving BERT Backbone from Scratch}}. In \bibinfo{booktitle}{\emph{AAAI Conference on Artificial Intelligence}}, Vol.~\bibinfo{volume}{36, no. 10}. \bibinfo{publisher}{AAAI}, \bibinfo{address}{Washington, DC, US}, \bibinfo{pages}{10663--10671}.
\newblock


\bibitem[\protect\citeauthoryear{Glickman}{Glickman}{2012}]%
        {glickman2012example}
\bibfield{author}{\bibinfo{person}{Mark~E Glickman}.} \bibinfo{year}{2012}\natexlab{}.
\newblock \bibinfo{booktitle}{\emph{{Example of the Glicko-2 system}}}.
\newblock \bibinfo{type}{{T}echnical {R}eport}. \bibinfo{institution}{Boston University}.
\newblock


\bibitem[\protect\citeauthoryear{Guo, Wang, Guo, Li, Song, Tan, Liu, Bian, and Yang}{Guo et~al\mbox{.}}{2024}]%
        {guo2024connecting}
\bibfield{author}{\bibinfo{person}{Qingyan Guo}, \bibinfo{person}{Rui Wang}, \bibinfo{person}{Junliang Guo}, \bibinfo{person}{Bei Li}, \bibinfo{person}{Kaitao Song}, \bibinfo{person}{Xu Tan}, \bibinfo{person}{Guoqing Liu}, \bibinfo{person}{Jiang Bian}, {and} \bibinfo{person}{Yujiu Yang}.} \bibinfo{year}{2024}\natexlab{}.
\newblock \bibinfo{title}{Connecting Large Language Models with Evolutionary Algorithms Yields Powerful Prompt Optimizers}.
\newblock
\newblock
\newblock
\shownote{arXiv:2309.08532}.


\bibitem[\protect\citeauthoryear{Hemberg, Moskal, and O'Reilly}{Hemberg et~al\mbox{.}}{2024}]%
        {hemberg2024evolving}
\bibfield{author}{\bibinfo{person}{Erik Hemberg}, \bibinfo{person}{Stephen Moskal}, {and} \bibinfo{person}{Una-May O'Reilly}.} \bibinfo{year}{2024}\natexlab{}.
\newblock \bibinfo{title}{Evolving Code with A Large Language Model}.
\newblock
\newblock
\newblock
\shownote{arXiv:2401.07102}.


\bibitem[\protect\citeauthoryear{Iacca, Caraffini, and Neri}{Iacca et~al\mbox{.}}{2015}]%
        {iacca2015continuous}
\bibfield{author}{\bibinfo{person}{Giovanni Iacca}, \bibinfo{person}{Fabio Caraffini}, {and} \bibinfo{person}{Ferrante Neri}.} \bibinfo{year}{2015}\natexlab{}.
\newblock \showarticletitle{Continuous Parameter Pools in Ensemble Self-Adaptive Differential Evolution}. In \bibinfo{booktitle}{\emph{IEEE Symposium Series on Computational Intelligence}}. \bibinfo{publisher}{{IEEE}}, \bibinfo{address}{{New York, NY, USA}}, \bibinfo{pages}{1529--1536}.
\newblock


\bibitem[\protect\citeauthoryear{Iacca, Neri, Caraffini, and Suganthan}{Iacca et~al\mbox{.}}{2014}]%
        {iacca2014differential}
\bibfield{author}{\bibinfo{person}{Giovanni Iacca}, \bibinfo{person}{Ferrante Neri}, \bibinfo{person}{Fabio Caraffini}, {and} \bibinfo{person}{Ponnuthurai~Nagaratnam Suganthan}.} \bibinfo{year}{2014}\natexlab{}.
\newblock \showarticletitle{A differential evolution framework with ensemble of parameters and strategies and pool of local search algorithms}. In \bibinfo{booktitle}{\emph{Applications of Evolutionary Computation: 17th European Conference, EvoApplications 2014, Granada, Spain, April 23-25, 2014, Revised Selected Papers 17}}. \bibinfo{publisher}{Springer}, \bibinfo{address}{Berlin Heidelberg, Germany}, \bibinfo{pages}{615--626}.
\newblock


\bibitem[\protect\citeauthoryear{{Jain, Naman and Vaidyanath, Skanda and Iyer, Arun and Natarajan, Nagarajan and Parthasarathy, Suresh and Rajamani, Sriram and Sharma, Rahul}}{{Jain, Naman and Vaidyanath, Skanda and Iyer, Arun and Natarajan, Nagarajan and Parthasarathy, Suresh and Rajamani, Sriram and Sharma, Rahul}}{2022}]%
        {10.1145/3510003.3510203}
\bibfield{author}{\bibinfo{person}{{Jain, Naman and Vaidyanath, Skanda and Iyer, Arun and Natarajan, Nagarajan and Parthasarathy, Suresh and Rajamani, Sriram and Sharma, Rahul}}.} \bibinfo{year}{{2022}}\natexlab{}.
\newblock \showarticletitle{{Jigsaw: Large Language Models Meet Program Synthesis}}. In \bibinfo{booktitle}{\emph{{International Conference on Software Engineering}}}. \bibinfo{publisher}{{Association for Computing Machinery}}, \bibinfo{address}{{New York, NY, USA}}, \bibinfo{pages}{{1219–1231}}.
\newblock
\showISBNx{{9781450392211}}


\bibitem[\protect\citeauthoryear{Jiang, Sablayrolles, Roux, Mensch, Savary, Bamford, Chaplot, de~las Casas, Hanna, Bressand, Lengyel, Bour, Lample, Lavaud, Saulnier, Lachaux, Stock, Subramanian, Yang, Antoniak, Scao, Gervet, Lavril, Wang, Lacroix, and Sayed}{Jiang et~al\mbox{.}}{2024}]%
        {jiang2024mixtral}
\bibfield{author}{\bibinfo{person}{Albert~Q. Jiang}, \bibinfo{person}{Alexandre Sablayrolles}, \bibinfo{person}{Antoine Roux}, \bibinfo{person}{Arthur Mensch}, \bibinfo{person}{Blanche Savary}, \bibinfo{person}{Chris Bamford}, \bibinfo{person}{Devendra~Singh Chaplot}, \bibinfo{person}{Diego de~las Casas}, \bibinfo{person}{Emma~Bou Hanna}, \bibinfo{person}{Florian Bressand}, \bibinfo{person}{Gianna Lengyel}, \bibinfo{person}{Guillaume Bour}, \bibinfo{person}{Guillaume Lample}, \bibinfo{person}{Lélio~Renard Lavaud}, \bibinfo{person}{Lucile Saulnier}, \bibinfo{person}{Marie-Anne Lachaux}, \bibinfo{person}{Pierre Stock}, \bibinfo{person}{Sandeep Subramanian}, \bibinfo{person}{Sophia Yang}, \bibinfo{person}{Szymon Antoniak}, \bibinfo{person}{Teven~Le Scao}, \bibinfo{person}{Théophile Gervet}, \bibinfo{person}{Thibaut Lavril}, \bibinfo{person}{Thomas Wang}, \bibinfo{person}{Timothée Lacroix}, {and} \bibinfo{person}{William~El Sayed}.} \bibinfo{year}{2024}\natexlab{}.
\newblock \bibinfo{title}{{Mixtral of Experts}}.
\newblock
\newblock
\newblock
\shownote{arXiv:2401.04088}.


\bibitem[\protect\citeauthoryear{{Jiao, Aoran and Patel, Tanmay P and Khurana, Sanjmi and Korol, Anna-Mariya and Brunke, Lukas and Adajania, Vivek K and Culha, Utku and Zhou, Siqi and Schoellig, Angela P}}{{Jiao, Aoran and Patel, Tanmay P and Khurana, Sanjmi and Korol, Anna-Mariya and Brunke, Lukas and Adajania, Vivek K and Culha, Utku and Zhou, Siqi and Schoellig, Angela P}}{2023}]%
        {jiao2023swarm}
\bibfield{author}{\bibinfo{person}{{Jiao, Aoran and Patel, Tanmay P and Khurana, Sanjmi and Korol, Anna-Mariya and Brunke, Lukas and Adajania, Vivek K and Culha, Utku and Zhou, Siqi and Schoellig, Angela P}}.} \bibinfo{year}{{2023}}\natexlab{}.
\newblock \bibinfo{title}{{Swarm-GPT: Combining Large Language Models with Safe Motion Planning for Robot Choreography Design}}.
\newblock
\newblock
\newblock
\shownote{{arXiv:2312.01059}}.


\bibitem[\protect\citeauthoryear{{Lanzi, Pier Luca and Loiacono, Daniele}}{{Lanzi, Pier Luca and Loiacono, Daniele}}{2023}]%
        {lanzi2023chatgpt}
\bibfield{author}{\bibinfo{person}{{Lanzi, Pier Luca and Loiacono, Daniele}}.} \bibinfo{year}{{2023}}\natexlab{}.
\newblock \bibinfo{title}{{ChatGPT and other large language models as evolutionary engines for online interactive collaborative game design}}.
\newblock
\newblock
\newblock
\shownote{{arXiv:2303.02155}}.


\bibitem[\protect\citeauthoryear{{Lehman, Joel and Gordon, Jonathan and Jain, Shawn and Ndousse, Kamal and Yeh, Cathy and Stanley, Kenneth O}}{{Lehman, Joel and Gordon, Jonathan and Jain, Shawn and Ndousse, Kamal and Yeh, Cathy and Stanley, Kenneth O}}{2023}]%
        {lehman2023evolution}
\bibfield{author}{\bibinfo{person}{{Lehman, Joel and Gordon, Jonathan and Jain, Shawn and Ndousse, Kamal and Yeh, Cathy and Stanley, Kenneth O}}.} \bibinfo{year}{{2023}}\natexlab{}.
\newblock \showarticletitle{{Evolution through large models}}.
\newblock In \bibinfo{booktitle}{\emph{{Handbook of Evolutionary Machine Learning}}}. \bibinfo{publisher}{{Springer}}, \bibinfo{address}{{Singapore}}, \bibinfo{pages}{{331--366}}.
\newblock


\bibitem[\protect\citeauthoryear{{Liu, Yang and Wu, Fanyou and Liu, Zhiyuan and Wang, Kai and Wang, Feiyue and Qu, Xiaobo}}{{Liu, Yang and Wu, Fanyou and Liu, Zhiyuan and Wang, Kai and Wang, Feiyue and Qu, Xiaobo}}{2023}]%
        {liu2023can}
\bibfield{author}{\bibinfo{person}{{Liu, Yang and Wu, Fanyou and Liu, Zhiyuan and Wang, Kai and Wang, Feiyue and Qu, Xiaobo}}.} \bibinfo{year}{{2023}}\natexlab{}.
\newblock \showarticletitle{{Can language models be used for real-world urban-delivery route optimization?}}
\newblock \bibinfo{journal}{\emph{{The Innovation}}} \bibinfo{volume}{{4}}, \bibinfo{number}{{6}} (\bibinfo{year}{{2023}}), \bibinfo{numpages}{{8}}~pages.
\newblock


\bibitem[\protect\citeauthoryear{{Michael R. Zhang and Nishkrit Desai and Juhan Bae and Jonathan Lorraine and Jimmy Ba}}{{Michael R. Zhang and Nishkrit Desai and Juhan Bae and Jonathan Lorraine and Jimmy Ba}}{2023}]%
        {zhang2023using}
\bibfield{author}{\bibinfo{person}{{Michael R. Zhang and Nishkrit Desai and Juhan Bae and Jonathan Lorraine and Jimmy Ba}}.} \bibinfo{year}{{2023}}\natexlab{}.
\newblock \bibinfo{title}{{Using Large Language Models for Hyperparameter Optimization}}.
\newblock
\newblock
\newblock
\shownote{{arXiv:2312.04528}}.


\bibitem[\protect\citeauthoryear{Pallagani, Roy, Muppasani, Fabiano, Loreggia, Murugesan, Srivastava, Rossi, Horesh, and Sheth}{Pallagani et~al\mbox{.}}{2024}]%
        {pallagani2024prospects}
\bibfield{author}{\bibinfo{person}{Vishal Pallagani}, \bibinfo{person}{Kaushik Roy}, \bibinfo{person}{Bharath Muppasani}, \bibinfo{person}{Francesco Fabiano}, \bibinfo{person}{Andrea Loreggia}, \bibinfo{person}{Keerthiram Murugesan}, \bibinfo{person}{Biplav Srivastava}, \bibinfo{person}{Francesca Rossi}, \bibinfo{person}{Lior Horesh}, {and} \bibinfo{person}{Amit Sheth}.} \bibinfo{year}{2024}\natexlab{}.
\newblock \bibinfo{title}{{On the Prospects of Incorporating Large Language Models (LLMs) in Automated Planning and Scheduling (APS)}}.
\newblock
\newblock
\newblock
\shownote{arXiv:2401.02500}.


\bibitem[\protect\citeauthoryear{{Pivithuru Thejan Amarasinghe and Su Nguyen and Yuan Sun and Damminda Alahakoon}}{{Pivithuru Thejan Amarasinghe and Su Nguyen and Yuan Sun and Damminda Alahakoon}}{2023}]%
        {amarasinghe2023aicopilot}
\bibfield{author}{\bibinfo{person}{{Pivithuru Thejan Amarasinghe and Su Nguyen and Yuan Sun and Damminda Alahakoon}}.} \bibinfo{year}{{2023}}\natexlab{}.
\newblock \bibinfo{title}{{AI-Copilot for Business Optimisation: A Framework and A Case Study in Production Scheduling}}.
\newblock
\newblock
\newblock
\shownote{{arXiv:2309.13218}}.


\bibitem[\protect\citeauthoryear{Prieto, Mengiste, and García~de Soto}{Prieto et~al\mbox{.}}{2023}]%
        {buildings13040857}
\bibfield{author}{\bibinfo{person}{Samuel~A. Prieto}, \bibinfo{person}{Eyob~T. Mengiste}, {and} \bibinfo{person}{Borja García~de Soto}.} \bibinfo{year}{2023}\natexlab{}.
\newblock \showarticletitle{{Investigating the Use of ChatGPT for the Scheduling of Construction Projects}}.
\newblock \bibinfo{journal}{\emph{Buildings}} \bibinfo{volume}{13}, \bibinfo{number}{4} (\bibinfo{year}{2023}), \bibinfo{numpages}{16}~pages.
\newblock
\showISSN{2075-5309}


\bibitem[\protect\citeauthoryear{Pryzant, Iter, Li, Lee, Zhu, and Zeng}{Pryzant et~al\mbox{.}}{2023}]%
        {pryzant2023automatic}
\bibfield{author}{\bibinfo{person}{Reid Pryzant}, \bibinfo{person}{Dan Iter}, \bibinfo{person}{Jerry Li}, \bibinfo{person}{Yin~Tat Lee}, \bibinfo{person}{Chenguang Zhu}, {and} \bibinfo{person}{Michael Zeng}.} \bibinfo{year}{2023}\natexlab{}.
\newblock \bibinfo{title}{{Automatic Prompt Optimization with ``Gradient Descent'' and Beam Search}}.
\newblock
\newblock
\newblock
\shownote{arXiv:2305.03495}.


\bibitem[\protect\citeauthoryear{{Qingyan Guo and Rui Wang and Junliang Guo and Bei Li and Kaitao Song and Xu Tan and Guoqing Liu and Jiang Bian and Yujiu Yang}}{{Qingyan Guo and Rui Wang and Junliang Guo and Bei Li and Kaitao Song and Xu Tan and Guoqing Liu and Jiang Bian and Yujiu Yang}}{2023}]%
        {guo2023connecting}
\bibfield{author}{\bibinfo{person}{{Qingyan Guo and Rui Wang and Junliang Guo and Bei Li and Kaitao Song and Xu Tan and Guoqing Liu and Jiang Bian and Yujiu Yang}}.} \bibinfo{year}{{2023}}\natexlab{}.
\newblock \bibinfo{title}{{Connecting Large Language Models with Evolutionary Algorithms Yields Powerful Prompt Optimizers}}.
\newblock
\newblock
\newblock
\shownote{{arXiv:2309.08532}}.


\bibitem[\protect\citeauthoryear{Rechenberg}{Rechenberg}{1973}]%
        {rechenberg1973evolutionsstrategie}
\bibfield{author}{\bibinfo{person}{Ingo Rechenberg}.} \bibinfo{year}{1973}\natexlab{}.
\newblock \bibinfo{booktitle}{\emph{{Evolutionsstrategie: Optimierung technischer Systeme nach Prinzipien derbiologischen Evolution}}}.
\newblock \bibinfo{publisher}{Frommann-Holzboog Verlag}, \bibinfo{address}{Stuttgart, Germany}.
\newblock


\bibitem[\protect\citeauthoryear{{Ren, Allen Z and Dixit, Anushri and Bodrova, Alexandra and Singh, Sumeet and Tu, Stephen and Brown, Noah and Xu, Peng and Takayama, Leila and Xia, Fei and Varley, Jake and others}}{{Ren, Allen Z and Dixit, Anushri and Bodrova, Alexandra and Singh, Sumeet and Tu, Stephen and Brown, Noah and Xu, Peng and Takayama, Leila and Xia, Fei and Varley, Jake and others}}{2023}]%
        {ren2023robots}
\bibfield{author}{\bibinfo{person}{{Ren, Allen Z and Dixit, Anushri and Bodrova, Alexandra and Singh, Sumeet and Tu, Stephen and Brown, Noah and Xu, Peng and Takayama, Leila and Xia, Fei and Varley, Jake and others}}.} \bibinfo{year}{{2023}}\natexlab{}.
\newblock \bibinfo{title}{{Robots that ask for help: Uncertainty alignment for large language model planners}}.
\newblock
\newblock
\newblock
\shownote{{arXiv:2307.01928}}.


\bibitem[\protect\citeauthoryear{{Romera-Paredes, Bernardino and Barekatain, Mohammadamin and Novikov, Alexander and Balog, Matej and Kumar, M Pawan and Dupont, Emilien and Ruiz, Francisco JR and Ellenberg, Jordan S and Wang, Pengming and Fawzi, Omar and others}}{{Romera-Paredes, Bernardino and Barekatain, Mohammadamin and Novikov, Alexander and Balog, Matej and Kumar, M Pawan and Dupont, Emilien and Ruiz, Francisco JR and Ellenberg, Jordan S and Wang, Pengming and Fawzi, Omar and others}}{2023}]%
        {romera2023mathematical}
\bibfield{author}{\bibinfo{person}{{Romera-Paredes, Bernardino and Barekatain, Mohammadamin and Novikov, Alexander and Balog, Matej and Kumar, M Pawan and Dupont, Emilien and Ruiz, Francisco JR and Ellenberg, Jordan S and Wang, Pengming and Fawzi, Omar and others}}.} \bibinfo{year}{{2023}}\natexlab{}.
\newblock \showarticletitle{{Mathematical discoveries from program search with large language models}}.
\newblock \bibinfo{journal}{\emph{{Nature}}}  \bibinfo{volume}{{(early access)}} (\bibinfo{year}{{2023}}), \bibinfo{pages}{{1--3}}.
\newblock


\bibitem[\protect\citeauthoryear{Seiler, Kerschke, and Trautmann}{Seiler et~al\mbox{.}}{2024}]%
        {seiler2024deepela}
\bibfield{author}{\bibinfo{person}{Moritz~Vinzent Seiler}, \bibinfo{person}{Pascal Kerschke}, {and} \bibinfo{person}{Heike Trautmann}.} \bibinfo{year}{2024}\natexlab{}.
\newblock \bibinfo{title}{Deep-ELA: Deep Exploratory Landscape Analysis with Self-Supervised Pretrained Transformers for Single- and Multi-Objective Continuous Optimization Problems}.
\newblock
\newblock
\newblock
\shownote{arXiv:2401.01192}.


\bibitem[\protect\citeauthoryear{Sun, Shao, Qian, Huang, and Qiu}{Sun et~al\mbox{.}}{2022}]%
        {pmlr-v162-sun22e}
\bibfield{author}{\bibinfo{person}{Tianxiang Sun}, \bibinfo{person}{Yunfan Shao}, \bibinfo{person}{Hong Qian}, \bibinfo{person}{Xuanjing Huang}, {and} \bibinfo{person}{Xipeng Qiu}.} \bibinfo{year}{2022}\natexlab{}.
\newblock \showarticletitle{{Black-Box Tuning for Language-Model-as-a-Service}}. In \bibinfo{booktitle}{\emph{International Conference on Machine Learning}} \emph{(\bibinfo{series}{Proceedings of Machine Learning Research}, Vol.~\bibinfo{volume}{162})}, \bibfield{editor}{\bibinfo{person}{Kamalika Chaudhuri}, \bibinfo{person}{Stefanie Jegelka}, \bibinfo{person}{Le~Song}, \bibinfo{person}{Csaba Szepesvari}, \bibinfo{person}{Gang Niu}, {and} \bibinfo{person}{Sivan Sabato}} (Eds.). \bibinfo{publisher}{PMLR}, \bibinfo{address}{Honolulu, HI, US}, \bibinfo{pages}{20841--20855}.
\newblock


\bibitem[\protect\citeauthoryear{{Sun, Lingfeng and Jha, Devesh K and Hori, Chiori and Jain, Siddarth and Corcodel, Radu and Zhu, Xinghao and Tomizuka, Masayoshi and Romeres, Diego}}{{Sun, Lingfeng and Jha, Devesh K and Hori, Chiori and Jain, Siddarth and Corcodel, Radu and Zhu, Xinghao and Tomizuka, Masayoshi and Romeres, Diego}}{2023}]%
        {sun2023interactive}
\bibfield{author}{\bibinfo{person}{{Sun, Lingfeng and Jha, Devesh K and Hori, Chiori and Jain, Siddarth and Corcodel, Radu and Zhu, Xinghao and Tomizuka, Masayoshi and Romeres, Diego}}.} \bibinfo{year}{{2023}}\natexlab{}.
\newblock \bibinfo{title}{{Interactive Planning Using Large Language Models for Partially Observable Robotics Tasks}}.
\newblock
\newblock
\newblock
\shownote{{arXiv:2312.06876}}.


\bibitem[\protect\citeauthoryear{{Tao, Ning and Ventresque, Anthony and Saber, Takfarinas}}{{Tao, Ning and Ventresque, Anthony and Saber, Takfarinas}}{2023}]%
        {tao2023program}
\bibfield{author}{\bibinfo{person}{{Tao, Ning and Ventresque, Anthony and Saber, Takfarinas}}.} \bibinfo{year}{{2023}}\natexlab{}.
\newblock \showarticletitle{{Program Synthesis with Generative Pre-trained Transformers and Grammar-Guided Genetic Programming Grammar}}. In \bibinfo{booktitle}{\emph{{IEEE Latin American Conference on Computational Intelligence}}}. \bibinfo{publisher}{{IEEE}}, \bibinfo{address}{{New York, NY, USA}}, \bibinfo{numpages}{{6}}~pages.
\newblock


\bibitem[\protect\citeauthoryear{Tessari and Iacca}{Tessari and Iacca}{2022}]%
        {tessari2022reinforcement}
\bibfield{author}{\bibinfo{person}{Michele Tessari} {and} \bibinfo{person}{Giovanni Iacca}.} \bibinfo{year}{2022}\natexlab{}.
\newblock \showarticletitle{Reinforcement learning based adaptive metaheuristics}. In \bibinfo{booktitle}{\emph{Genetic and Evolutionary Computation Conference Companion}}. \bibinfo{publisher}{{ACM}}, \bibinfo{address}{{New York, NY, USA}}, \bibinfo{pages}{1854--1861}.
\newblock


\bibitem[\protect\citeauthoryear{Touvron, Martin, Stone, Albert, Almahairi, Babaei, Bashlykov, Batra, Bhargava, Bhosale, Bikel, Blecher, Ferrer, Chen, Cucurull, Esiobu, Fernandes, Fu, Fu, Fuller, Gao, Goswami, Goyal, Hartshorn, Hosseini, Hou, Inan, Kardas, Kerkez, Khabsa, Kloumann, Korenev, Koura, Lachaux, Lavril, Lee, Liskovich, Lu, Mao, Martinet, Mihaylov, Mishra, Molybog, Nie, Poulton, Reizenstein, Rungta, Saladi, Schelten, Silva, Smith, Subramanian, Tan, Tang, Taylor, Williams, Kuan, Xu, Yan, Zarov, Zhang, Fan, Kambadur, Narang, Rodriguez, Stojnic, Edunov, and Scialom}{Touvron et~al\mbox{.}}{2023}]%
        {touvron2023llama}
\bibfield{author}{\bibinfo{person}{Hugo Touvron}, \bibinfo{person}{Louis Martin}, \bibinfo{person}{Kevin Stone}, \bibinfo{person}{Peter Albert}, \bibinfo{person}{Amjad Almahairi}, \bibinfo{person}{Yasmine Babaei}, \bibinfo{person}{Nikolay Bashlykov}, \bibinfo{person}{Soumya Batra}, \bibinfo{person}{Prajjwal Bhargava}, \bibinfo{person}{Shruti Bhosale}, \bibinfo{person}{Dan Bikel}, \bibinfo{person}{Lukas Blecher}, \bibinfo{person}{Cristian~Canton Ferrer}, \bibinfo{person}{Moya Chen}, \bibinfo{person}{Guillem Cucurull}, \bibinfo{person}{David Esiobu}, \bibinfo{person}{Jude Fernandes}, \bibinfo{person}{Jeremy Fu}, \bibinfo{person}{Wenyin Fu}, \bibinfo{person}{Brian Fuller}, \bibinfo{person}{Cynthia Gao}, \bibinfo{person}{Vedanuj Goswami}, \bibinfo{person}{Naman Goyal}, \bibinfo{person}{Anthony Hartshorn}, \bibinfo{person}{Saghar Hosseini}, \bibinfo{person}{Rui Hou}, \bibinfo{person}{Hakan Inan}, \bibinfo{person}{Marcin Kardas}, \bibinfo{person}{Viktor Kerkez}, \bibinfo{person}{Madian Khabsa},
  \bibinfo{person}{Isabel Kloumann}, \bibinfo{person}{Artem Korenev}, \bibinfo{person}{Punit~Singh Koura}, \bibinfo{person}{Marie-Anne Lachaux}, \bibinfo{person}{Thibaut Lavril}, \bibinfo{person}{Jenya Lee}, \bibinfo{person}{Diana Liskovich}, \bibinfo{person}{Yinghai Lu}, \bibinfo{person}{Yuning Mao}, \bibinfo{person}{Xavier Martinet}, \bibinfo{person}{Todor Mihaylov}, \bibinfo{person}{Pushkar Mishra}, \bibinfo{person}{Igor Molybog}, \bibinfo{person}{Yixin Nie}, \bibinfo{person}{Andrew Poulton}, \bibinfo{person}{Jeremy Reizenstein}, \bibinfo{person}{Rashi Rungta}, \bibinfo{person}{Kalyan Saladi}, \bibinfo{person}{Alan Schelten}, \bibinfo{person}{Ruan Silva}, \bibinfo{person}{Eric~Michael Smith}, \bibinfo{person}{Ranjan Subramanian}, \bibinfo{person}{Xiaoqing~Ellen Tan}, \bibinfo{person}{Binh Tang}, \bibinfo{person}{Ross Taylor}, \bibinfo{person}{Adina Williams}, \bibinfo{person}{Jian~Xiang Kuan}, \bibinfo{person}{Puxin Xu}, \bibinfo{person}{Zheng Yan}, \bibinfo{person}{Iliyan Zarov}, \bibinfo{person}{Yuchen
  Zhang}, \bibinfo{person}{Angela Fan}, \bibinfo{person}{Melanie Kambadur}, \bibinfo{person}{Sharan Narang}, \bibinfo{person}{Aurelien Rodriguez}, \bibinfo{person}{Robert Stojnic}, \bibinfo{person}{Sergey Edunov}, {and} \bibinfo{person}{Thomas Scialom}.} \bibinfo{year}{2023}\natexlab{}.
\newblock \bibinfo{title}{{Llama 2: Open Foundation and Fine-Tuned Chat Models}}.
\newblock
\newblock
\newblock
\shownote{arXiv:2307.09288}.


\bibitem[\protect\citeauthoryear{Tribes, Benarroch-Lelong, Lu, and Kobyzev}{Tribes et~al\mbox{.}}{2024}]%
        {tribes2024hyperparameter}
\bibfield{author}{\bibinfo{person}{Christophe Tribes}, \bibinfo{person}{Sacha Benarroch-Lelong}, \bibinfo{person}{Peng Lu}, {and} \bibinfo{person}{Ivan Kobyzev}.} \bibinfo{year}{2024}\natexlab{}.
\newblock \bibinfo{title}{{Hyperparameter Optimization for Large Language Model Instruction-Tuning}}.
\newblock
\newblock
\newblock
\shownote{arXiv:2312.00949}.


\bibitem[\protect\citeauthoryear{{Vo{\ss}, Stefan}}{{Vo{\ss}, Stefan}}{2023}]%
        {chatGPTInLogistics}
\bibfield{author}{\bibinfo{person}{{Vo{\ss}, Stefan}}.} \bibinfo{year}{{2023}}\natexlab{}.
\newblock \showarticletitle{{Successfully Using ChatGPT in Logistics: Are We There Yet?}}. In \bibinfo{booktitle}{\emph{{Computational Logistics}}}. \bibinfo{publisher}{{Springer Nature}}, \bibinfo{address}{{Cham, Switzerland}}, \bibinfo{pages}{{3--17}}.
\newblock


\bibitem[\protect\citeauthoryear{{Wang, Lirui and Ling, Yiyang and Yuan, Zhecheng and Shridhar, Mohit and Bao, Chen and Qin, Yuzhe and Wang, Bailin and Xu, Huazhe and Wang, Xiaolong}}{{Wang, Lirui and Ling, Yiyang and Yuan, Zhecheng and Shridhar, Mohit and Bao, Chen and Qin, Yuzhe and Wang, Bailin and Xu, Huazhe and Wang, Xiaolong}}{2023}]%
        {wang2023gensim}
\bibfield{author}{\bibinfo{person}{{Wang, Lirui and Ling, Yiyang and Yuan, Zhecheng and Shridhar, Mohit and Bao, Chen and Qin, Yuzhe and Wang, Bailin and Xu, Huazhe and Wang, Xiaolong}}.} \bibinfo{year}{{2023}}\natexlab{}.
\newblock \bibinfo{title}{{GenSim: Generating Robotic Simulation Tasks via Large Language Models}}.
\newblock
\newblock
\newblock
\shownote{{arXiv:2310.01361}}.


\bibitem[\protect\citeauthoryear{Wu, hao Wu, Wu, Feng, and Tan}{Wu et~al\mbox{.}}{2024}]%
        {wu2024evolutionary}
\bibfield{author}{\bibinfo{person}{Xingyu Wu}, \bibinfo{person}{Sheng hao Wu}, \bibinfo{person}{Jibin Wu}, \bibinfo{person}{Liang Feng}, {and} \bibinfo{person}{Kay~Chen Tan}.} \bibinfo{year}{2024}\natexlab{}.
\newblock \bibinfo{title}{{Evolutionary Computation in the Era of Large Language Model: Survey and Roadmap}}.
\newblock
\newblock
\newblock
\shownote{arXiv:2401.10034}.


\bibitem[\protect\citeauthoryear{{Xianggen Liu and Pengyong Li and Fandong Meng and Hao Zhou and Huasong Zhong and Jie Zhou and Lili Mou and Sen Song}}{{Xianggen Liu and Pengyong Li and Fandong Meng and Hao Zhou and Huasong Zhong and Jie Zhou and Lili Mou and Sen Song}}{2021}]%
        {LIU2021310}
\bibfield{author}{\bibinfo{person}{{Xianggen Liu and Pengyong Li and Fandong Meng and Hao Zhou and Huasong Zhong and Jie Zhou and Lili Mou and Sen Song}}.} \bibinfo{year}{{2021}}\natexlab{}.
\newblock \showarticletitle{{Simulated annealing for optimization of graphs and sequences}}.
\newblock \bibinfo{journal}{\emph{{Neurocomputing}}}  \bibinfo{volume}{{465}} (\bibinfo{year}{{2021}}), \bibinfo{pages}{{310--324}}.
\newblock
\showISSN{{0925-2312}}


\bibitem[\protect\citeauthoryear{{Yaman, Anil and Hallawa, Ahmed and Coler, Matt and Iacca, Giovanni}}{{Yaman, Anil and Hallawa, Ahmed and Coler, Matt and Iacca, Giovanni}}{2017}]%
        {yaman2017presenting}
\bibfield{author}{\bibinfo{person}{{Yaman, Anil and Hallawa, Ahmed and Coler, Matt and Iacca, Giovanni}}.} \bibinfo{year}{{2017}}\natexlab{}.
\newblock \showarticletitle{{Presenting the ECO: evolutionary computation ontology}}. In \bibinfo{booktitle}{\emph{{Applications of Evolutionary Computation}}}. \bibinfo{publisher}{{Springer}}, \bibinfo{address}{{Cham, Switzerland}}, \bibinfo{pages}{{603--619}}.
\newblock


\bibitem[\protect\citeauthoryear{{Yaman, Anil and Iacca, Giovanni and Caraffini, Fabio}}{{Yaman, Anil and Iacca, Giovanni and Caraffini, Fabio}}{2019}]%
        {yaman2019comparison}
\bibfield{author}{\bibinfo{person}{{Yaman, Anil and Iacca, Giovanni and Caraffini, Fabio}}.} \bibinfo{year}{{2019}}\natexlab{}.
\newblock \showarticletitle{{A comparison of three differential evolution strategies in terms of early convergence with different population sizes}}. In \bibinfo{booktitle}{\emph{{AIP Conference Proceedings}}}, Vol.~\bibinfo{volume}{{2070}}. \bibinfo{publisher}{{AIP Publishing}}, \bibinfo{address}{{Melville, NY, USA}}, \bibinfo{numpages}{{4}}~pages.
\newblock


\bibitem[\protect\citeauthoryear{Zhang, Gong, Wu, Liu, and Zhou}{Zhang et~al\mbox{.}}{2023}]%
        {zhang2023automlgpt}
\bibfield{author}{\bibinfo{person}{Shujian Zhang}, \bibinfo{person}{Chengyue Gong}, \bibinfo{person}{Lemeng Wu}, \bibinfo{person}{Xingchao Liu}, {and} \bibinfo{person}{Mingyuan Zhou}.} \bibinfo{year}{2023}\natexlab{}.
\newblock \bibinfo{title}{{AutoML-GPT: Automatic Machine Learning with GPT}}.
\newblock
\newblock
\newblock
\shownote{arXiv:2305.02499}.


\bibitem[\protect\citeauthoryear{Zhao, Wang, and Yang}{Zhao et~al\mbox{.}}{2023}]%
        {ijcai2023p588}
\bibfield{author}{\bibinfo{person}{Jiangjiang Zhao}, \bibinfo{person}{Zhuoran Wang}, {and} \bibinfo{person}{Fangchun Yang}.} \bibinfo{year}{2023}\natexlab{}.
\newblock \showarticletitle{{Genetic Prompt Search via Exploiting Language Model Probabilities}}. In \bibinfo{booktitle}{\emph{International Joint Conference on Artificial Intelligence}}, \bibfield{editor}{\bibinfo{person}{Edith Elkind}} (Ed.). \bibinfo{publisher}{International Joint Conferences on Artificial Intelligence Organization}, \bibinfo{address}{Macao}, \bibinfo{pages}{5296--5305}.
\newblock


\bibitem[\protect\citeauthoryear{Zheng, Su, You, Wang, Qian, Xu, and Albanie}{Zheng et~al\mbox{.}}{2023}]%
        {zheng2023gpt4}
\bibfield{author}{\bibinfo{person}{Mingkai Zheng}, \bibinfo{person}{Xiu Su}, \bibinfo{person}{Shan You}, \bibinfo{person}{Fei Wang}, \bibinfo{person}{Chen Qian}, \bibinfo{person}{Chang Xu}, {and} \bibinfo{person}{Samuel Albanie}.} \bibinfo{year}{2023}\natexlab{}.
\newblock \bibinfo{title}{{Can GPT-4 Perform Neural Architecture Search?}}
\newblock
\newblock
\newblock
\shownote{arXiv:2304.10970}.


\end{thebibliography}

\end{document}